\documentclass[conference]{IEEEtran}
\IEEEoverridecommandlockouts
\usepackage{cite}
\usepackage{amsmath,amssymb,amsfonts}
\usepackage{algorithm}
\usepackage{algorithmicx}
\usepackage{algpseudocode}
\usepackage{graphicx}
\usepackage{textcomp}
\usepackage{xcolor}
\usepackage{booktabs}
\usepackage[
    detect-all=true,
    detect-family=true,
    group-digits=false,
    separate-uncertainty=true
]{siunitx}
\usepackage{cleveref}
\usepackage{multirow}
\usepackage{adjustbox}
\usepackage{tikz}
\usepackage{bm}
\usepackage[letterpaper]{geometry}
\def\BibTeX{{\rm B\kern-.05em{\sc i\kern-.025em b}\kern-.08em
    T\kern-.1667em\lower.7ex\hbox{E}\kern-.125emX}}


    
\begin{document}
\newgeometry{top=1in, bottom=0.75in, left=0.75in, right=0.75in}
\title{Diffusion Models for Safety Validation of Autonomous Driving Systems}
\author{\IEEEauthorblockN{Juanran Wang\textsuperscript{*}}
\IEEEauthorblockA{\textit{Computer Science} \\
\textit{Stanford University}\\
Stanford, CA, USA \\
jun2026@stanford.edu}
\and
\IEEEauthorblockN{Marc R. Schlichting\textsuperscript{*}}
\IEEEauthorblockA{\textit{Aeronautics and Astronautics} \\
\textit{Stanford University}\\
Stanford, CA, USA \\
mschl@stanford.edu}
\and

\IEEEauthorblockN{Harrison Delecki}
\IEEEauthorblockA{\textit{Aeronautics and Astronautics} \\
\textit{Stanford University}\\
Stanford, CA, USA \\
hdelecki@stanford.edu}
\and

\IEEEauthorblockN{Mykel J. Kochenderfer}
\IEEEauthorblockA{\textit{Aeronautics and Astronautics} \\
\textit{Stanford University}\\
Stanford, CA, USA \\
mykel@stanford.edu}
}

\maketitle

\begingroup
\renewcommand\thefootnote{\textsuperscript{*}}
\footnotetext{These authors contributed equally to this work.}
\endgroup

\begin{abstract}
Safety validation of autonomous driving systems is extremely challenging due to the high risks and costs of real-world testing as well as the rarity and diversity of potential failures. To address these challenges, we train a denoising diffusion model to generate potential failure cases of an autonomous vehicle given any initial traffic state. Experiments on a four-way intersection problem show that in a variety of scenarios, the diffusion model can generate realistic failure samples while capturing a wide variety of potential failures. Our model does not require any external training dataset, can perform training and inference with modest computing resources, and does not assume any prior knowledge of the system under test, with applicability to safety validation for traffic intersections.
\end{abstract}

\section{Introduction}
\label{intro}
Verifying the safety of an autonomous driving system requires knowledge of the potential failure modes of the system. Learning the failure distribution of the system under test (SUT) presents a wide variety of challenges including:
\begin{itemize}
    \item \textit{Curse of dimensionality}: The high-dimensional nature of the state space of the autonomous vehicle and the long time horizons over which it operates.
    \item \textit{Curse of sparsity}: Failures tend to be rare for most safety-critical autonomous driving systems, which makes the search for failures highly time-consuming.
    \item \textit{Curse of multimodality}: The system may exhibit multiple failure modes of different nature, making it hard to capture the full range of variability in its failure distribution.
\end{itemize}

Due to the high risks and costs of deploying the system on hardware for scanerio-based and real-world testing \cite{junietz2018evaluation}, there has been growing interest in designing effective virtual testing frameworks. Most of these frameworks typically involve constructing a virtual simulation environment and running pre-designed simulated tests tailored to the system\cite{dona2022virtual}. These testing frameworks, however, often rely on extensive prior knowledge of the SUT and require significant redesigns when being adapted to validate different driving systems.

Instead of gathering as many individual failure cases as possible through simulation or real-world testing, directly modeling the true failure distribution of the SUT can provide more comprehensive insights into the failure modes of the system. Recently, denoising diffusion models have demonstrated the capacity for generative modeling of complex target distributions, enabling the synthesis of high-quality texts, images, and robotic actions \cite{saharia2022photorealistic, rombach2022high, janner2022planning}.

\begin{figure}[t]
  \includegraphics[width=\linewidth]{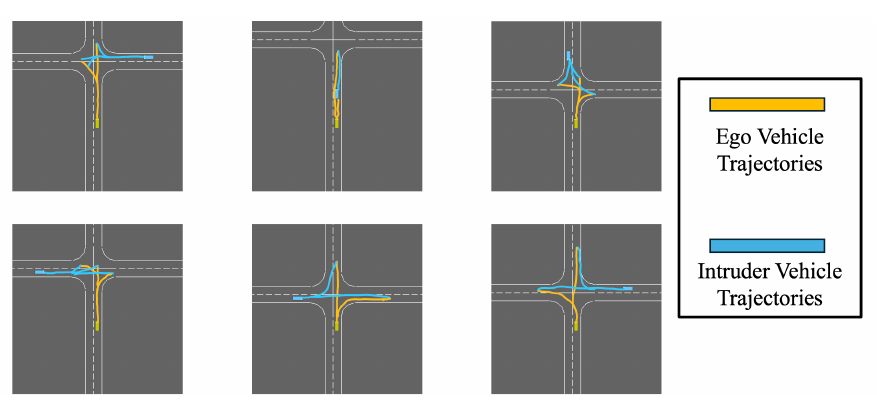}
  \caption{Possible collision trajectories for different initial states of traffic intersection. In this study, we test our diffusion-based safety validation framework on the traffic intersection problem, wherein the ego vehicle attempts to navigate the intersection in the presence of an intruder. We train a diffusion model to generate possible collision trajectories of the vehicles based on any initial state of the intersection.}
  \label{overall viz}
\end{figure}

This study applies diffusion models to perform safety validation for autonomous vehicles. Given an autonomous driving system, we employ a diffusion model to learn the true distribution over collision-causing sequences of observation errors of the SUT under different traffic situations. Our training pipeline does not require any external dataset. At each iteration, the model generates a batch of failure-causing sensor disturbance samples and trains itself on samples closer to actual collisions. During inference, the model can generate failure samples based on any given initial traffic state. As shown in \cref{overall viz}, to test our framework, we train the diffusion model to generate failure samples for a four-way intersection problem involving an autonomous ego vehicle and one intruder vehicle. The simulation environment randomly initializes the positions, velocities, and routes of the vehicles.

%\vspace{-0.1cm}
Our code is pubicly available.\footnote{https://github.com/sisl/Diffusion-Based-AV-Safety-Validation} Our main contribution is a framework with the following properties:
\newgeometry{top=0.7in, bottom=0.7in, left=0.7in, right=0.7in}
\begin{itemize}
    \item It supports the generation of failure-causing sensor noise samples \textit{based on any specified initial state of the intersection}.
    \item It generates \textit{realistic} sensor noise samples while \textit{covering a wide variety of possible failures} in many traffic scenarios.
    \item It is substantially \textit{more sample-efficient} in generating collision-causing sensor noise than Monte Carlo simulation methods.
    \item It uses a multi-stage self-improvement training strategy with \textit{no need for an external dataset}.
    \item It does not assume any prior knowledge of the SUT.
    \item It supports training and inference on modest hardware.
\end{itemize}

\Cref{Related Work} discusses the current state of autonomous vehicle safety validation research. \Cref{Methods} formally defines our problem, introduces the architecture of diffusion models, presents our training algorithm, and explains our evaluation metrics and baselines. \Cref{Experiments} and \ref{Discussions} presents, visualizes, and discusses our experimental results.

\section{Related Work}\label{Related Work}
Early attempts at computational safety validation of autonomous driving systems involve constructing specialized simulation environments designed for the SUT and performing a large number of simulations to identify possible failure modes \cite{dona2022virtual, li2023simulation, tlig2018autonomous}. However, these testing frameworks are often tailored to the SUT and may not be applicable to the validation of different system designs. Some statistical approaches create mathematical models of the risk space from real-world driving data or prior knowledge. They then perform importance sampling to generate failure samples or statistical inference to estimate the failure rate\cite{jiang2022efficient, xu2019quantitative}. These strategies, however, require extensive prior knowledge or data about the real-world operational experience of the SUT, which may not be readily available for newly designed systems. Formal verification-based methods such as Interpretable Validation find temporal logic expressions that describe the most likely failures of the SUT. While offering greater interpretability, this framework generally fails to identify a diverse set of potential failure modes as it focuses on the most likely failures of the system\cite{corso2020interpretable}. Finally, reinforcement learning-based frameworks train an adversarial agent to intentionally produce collision scenarios in order to identify potential failures \cite{feng2023dense, koren2018adaptive}. Although these methods allow for the generation of failure scenarios based on a particular initial state of interest, they often require a large amount of data or extensive simulation runs.

Over the past few years, diffusion models have shown substantial improvement in their ability to synthesize high-quality data in various fields\cite{ho2020denoising,rombach2022high,saharia2022photorealistic}. In particular, Denoising Diffusion Probabilistic Models (DDPMs) have demonstrated strong performance in generative modeling of time-series data, maintaining fidelity to the overall data distribution while preserving the distribution of extreme values during generation\cite{galib2024fide}. DDPMs are thus a promising method for generative modeling of temporal failure trajectories in safety-critical systems, where failures may be rare and yet costly. 

Recent frameworks such as AdvDiffuser and DrivingGen use real-world driving logs or video recordings to train a diffusion model for failure generation \cite{chen2023advdiffuser,guo2024drivinggen}. However, these methods require a large number of real-world failure records for training, which may be hard to collect especially for novel driving systems. Other methods including SAFE-SIM employ guided diffusion models to synthesize aggressive actions for a chosen adversarial vehicle in order to stress-test the ego vehicle\cite{chang2025safe}. While SAFE-SIM allows users to specify key aspects of the scenario during generation, it still requires rule-based proposal trajectories for the adversarial vehicle which must be designed with prior knowledge of the SUT, limiting its generalizability. Finally, frameworks like DiFS can perform diffusion-based failure sampling for any cyber-physical system in general; however, DiFS assumes that the initial state of the problem is fixed, which limits its ability to model the full failure distribution of the system in real-world conditions\cite{delecki2024diffusion}. This study addresses these challenges by training a diffusion model to generate failure samples of an autonomous driving system based on any given initial traffic state. Our framework does not require any external training dataset and does not assume any prior knowledge of the SUT.

\section{Methods} \label{Methods}
\subsection{Problem Formulation}\label{setup}
\begin{figure}[b]
  % \begin{adjustbox}{max width=\columnwidth}
  \includegraphics[width=\columnwidth]{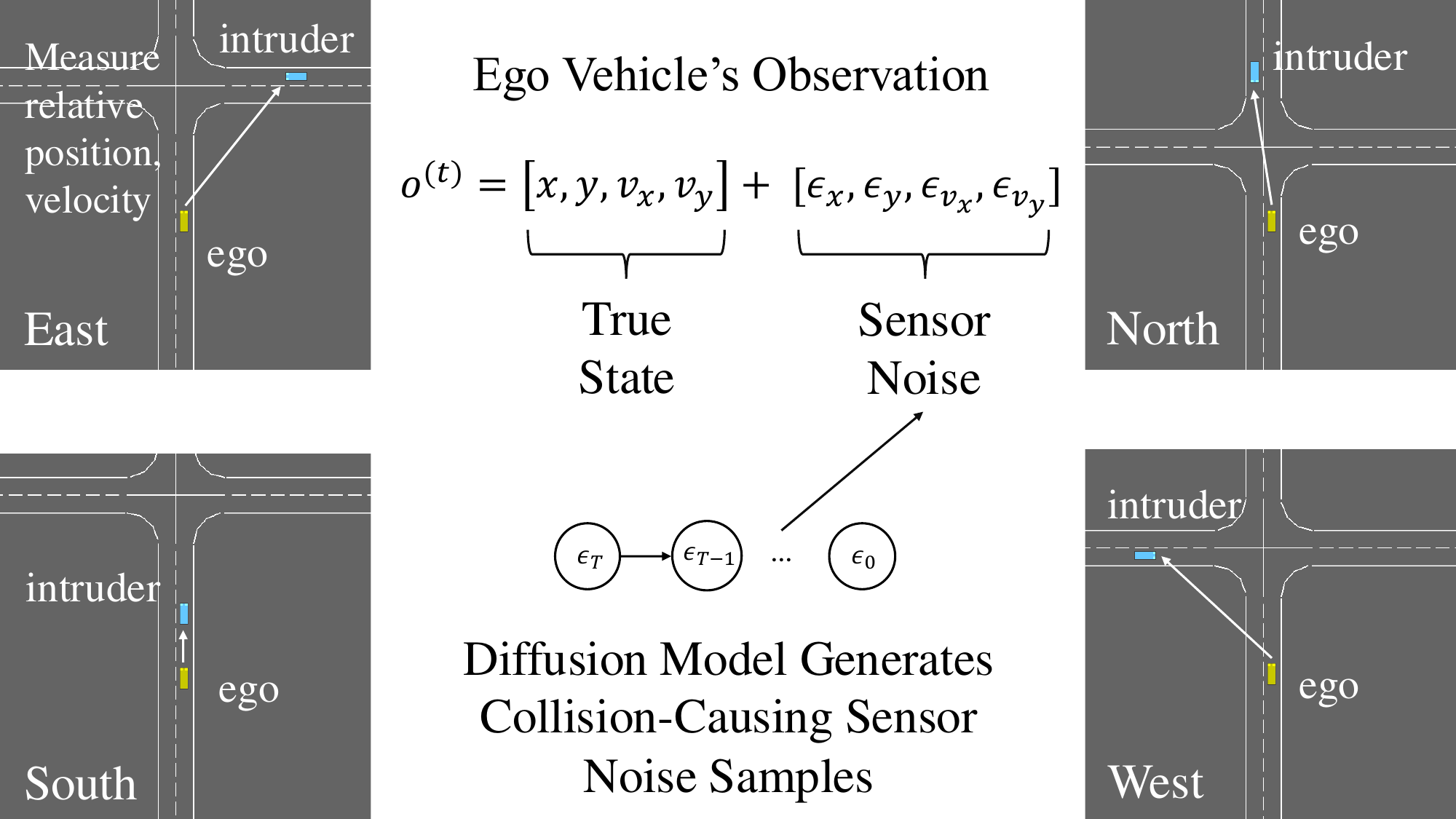}
  \vspace{-0.5cm}
  \caption{Illustration of our problem setup. A self-driving ego vehicle attempts to navigate an intersection in the presence of an intruder vehicle, which may approach the intersection from the east, west, south, and north road branch. Our goal is to train a diffusion model to generate collision-causing sensor noise samples.}
  \label{problem viz}
  % \end{adjustbox}
\end{figure}

This study focuses on sampling potential failures of an autonomous vehicle navigating a four-way intersection, using HighwayEnv's intersection environment for simulations.\footnote{https://github.com/Farama-Foundation/HighwayEnv} As shown in \cref{problem viz}, we consider an intersection with four road branches: south (bottom), north (top), west (left), and east (right).  Each has two lanes for the two directions of travel. The ego vehicle emerges from the south branch and must navigate the intersection to reach one of the other three branches in the presence of an intruder vehicle. The intruder vehicle might emerge from any of the four branches and might turn left, go straight, or turn right at the intersection. The initial positions and velocities of the two vehicles are randomized. Both vehicles are longitudinally controlled by the Intelligent Driver Model \cite{kesting2010enhanced} within lanes. The ego vehicle follows a fixed policy, while the intruder vehicle is configured to exhibit randomized behavior by randomly sampling the delta-exponent of the IDM, using HighwayEnv's provided feature.

The ego vehicle can observe the position and velocity of the intruder vehicle relative to itself $x, y, v_x, v_y$ through a noisy sensor. We assume that the ego vehicle does not know the intruder's destination road branch, and the prior sensor noise model is a Gaussian distribution, $\epsilon^{(t)} = [\epsilon_x, \epsilon_y, \epsilon_{v_x}, \epsilon_{v_y}] \sim \mathcal{N}(\mathbf{0}, \gamma \mathbf{I})$, where $\gamma$ is the noise scale. Based on each observation, the ego vehicle performs an action, which corresponds to the acceleration along its planned path of motion (which it does not modify) over the next time step. We consider a horizon of 24 timesteps: the ego vehicle performs 23 observations and actions starting from the initial timestep to complete a \textit{simulation}. The \textit{robustness} $\rho$ of a simulation is defined as the minimum separating distance between the vehicles throughout the simulation. A $\rho$ of 0 means that a collision occurs and thus the simulation is a \textit{failure}. We define a \textit{scenario} $\mathbf{S}$ as the set of possible simulations with the intruder emerging from a specific road branch. Since the intruder may emerge from any of the four branches (south, north, east, or west), there are four scenarios in total. In each scenario, given the initial relative position and velocity of the vehicles $\textbf{s}_{0} = [x_0, y_0, v_{x, 0}, v_{y, 0}]$, and a particular temporal series of the ego vehicle's observation error $\bm{\epsilon} = [\epsilon^{(t=1)}, \dots, \epsilon^{(t=23)}]$, a set of different simulations may occur depending on the randomized behavior of the intruder vehicle and the destination of the ego vehicle. The robustness of a simulation is therefore a random variable conditioned on the initial state and the observation error, $\rho_{\mathbf{S}}(\textbf{s}_{0}, \bm{\epsilon})$. The problem of sampling potential failures of the ego vehicle can thus be formulated as follows:

\textbf{Problem.} For each scenario $\mathbf{S} \in \{\text{east}, \text{west}, \text{south}, \text{north}\}$, given the initial relative position and velocity of the ego and intruder, $\textbf{s}_{0}$, we want to sample temporal series of observation error $\bm{\epsilon}$ from:
\begin{equation}\label{targetdistribution}
    p_{\mathbf{S}}(\bm{\epsilon} \mid \rho=0, \textbf{s}_{0}) \propto p(\rho_{\mathbf{S}}(\textbf{s}_{0}, \bm{\epsilon})=0)p(\bm{\epsilon})
\end{equation}
where $p(\bm{\epsilon})=\mathcal{N}(\bm{\epsilon}\mid \mathbf{0}, \gamma \mathbf{I})$ assuming Gaussian sensor noise. This target distribution is difficult to represent analytically; therefore, we propose to train a deep generative model to generate high-fidelity samples from the distribution.

\subsection{Conditional Diffusion Sampling}
Recent studies \cite{galib2024fide} have demonstrated the effectiveness of DDPMs in generating time series data while capturing the distribution of abnormal and extreme values for risk management applications. We apply conditional DDPMs \cite{ho2020denoising} to the problem of generating time series of observation error that likely cause the ego vehicle to collide with the intruder. The diffusion model learns to transform a Gaussian distribution to the target distribution over observation errors as in \cref{targetdistribution} through a series of $K$ denoising steps. For a particular scenario and initial state $\mathbf{s}_0$, let $\bm{\epsilon}_0$ be a sample from our target distribution. The forward diffusion process gradually adds noise to $\bm{\epsilon}_0$ over $K$ steps, 
\begin{equation}\label{forwarddiffusion}
\bm{\epsilon}_k \sim \mathcal{N}(\sqrt{1-\beta_k}\epsilon_{k-1}, \beta_k \mathbf{I}), k=1, \dots, K
\end{equation} which transforms $\bm{\epsilon}_0$ to a unit Gaussian at the final step. The variance schedule $\beta_1, \dots, \beta_k$ is a fixed increasing sequence. Then, for the reverse diffusion process, the model is trained to reverse the noise-adding forward process, transforming a unit Gaussian $\bm{\epsilon}_K$ to a sample in our target distribution $\mathbf{\epsilon_0}$,
\begin{equation}\label{reversediffusion}
\bm{\epsilon}_{k-1} \sim \mathcal{N}(\mathbf{\mu}_{\theta}(\bm{\epsilon}_{k} \mid k, \rho_{\text{threshold}}, \mathbf{s}_0), \beta_k \mathbf{I}), k=K, \dots, 1
\end{equation} thus generating a sample in the target distribution. We use a U-Net \cite{ronneberger2015u} neural network model to approximate $\mathbf{\mu}_{\theta}$, predicting the mean of $\bm{\epsilon}_{k-1}$ from the data after the previous denoising step $\bm{\epsilon}_{k}$, conditioned on the denoising step $k$, the initial state of the vehicles $\mathbf{s}_0$, and the robustness threshold $\rho_{\text{threshold}}$, which specifies the desired robustness of simulations with the generated observation error sample. Our diffusion model samples from the distribution with density $p_{\theta}(\bm{\epsilon}_{0:K} \mid \rho_{\text{threshold}}, \mathbf{s}_0)$, which is given by 
\begin{equation} \label{diffusionmodel}
    p(\bm{\epsilon}_K)\prod_{k=1}^{K} \mathcal{N}(\bm{\epsilon}_{k-1} \mid \mathbf{\mu}_{\theta}(\bm{\epsilon}_{k} \mid k, \rho_{\text{threshold}}, \mathbf{s}_0), \beta_k \mathbf{I})
\end{equation}
where $p(\bm{\epsilon}_K) = \mathcal{N}(\bm{\epsilon}_K \mid \mathbf{0}, \mathbf{I})$ because the initial noise is Gaussian. During inference, we set $\rho_{\text{threshold}}=0$ since we want the model to generate observation error samples that lead to collisions $(\rho=0)$. During training, as explained in the next section, we may set $\rho_{\text{threshold}}$ to different values, guiding the diffusion model to progressively generate observation error samples that lead to simulations of lower and lower robustness.

\subsection{Training Algorithm}
Due to the sparsity and multimodality of failures in our problem as mentioned in \cref{intro}, it may be challenging to directly train the diffusion model to generate collision-causing observation error samples. Hence, we propose a multi-stage training algorithm (\cref{alg:diffusion_failure_sampling}) which has the diffusion model perform inference with progressively lower robustness thresholds. This guides the model to generate samples that result in simulations closer and closer to collision. The main steps of our algorithm are as follows:

\begin{itemize}
    \item Reinitialize the simulation environments.
    \item The diffusion model generates samples based on the initial vehicle states and robustness thresholds sampled uniformly between 0 and $\Tilde{\rho}$ (the robustness cutoff for elite samples). Append the generated samples to the dataset.
    \item Perform the simulations with the generated samples to obtain their robustness values. Reset $\Tilde{\rho}$ to the $\alpha$-th percentile robustness value of the generated samples.
    \item Select elite samples from the dataset with robustness lower than $\Tilde{\rho}$ and train the diffusion model with these elite samples \cite{ho2020denoising}.
    \item Repeat the above steps until $\Tilde{\rho}$ converges or reaches 0.
\end{itemize}

At each stage, we query the model with robustness thresholds that fall within the lowest $\alpha$ robustness range of the samples generated in the previous stage. Therefore, we can expect the samples generated at each stage to cause lower-robustness simulations than those generated at the previous stage. The model thus generates and gets trained on samples with decreasing robustness over time. This helps concentrate the training effort on low-robustness samples in the later stages, making the model more effective in generating observation errors that lead to collisions or near-collisions.

\begin{algorithm}[t]
\caption{Training Diffusion Model for Failure Generation}
\label{alg:diffusion_failure_sampling}
\begin{algorithmic}[1] % [1] ensures line numbering
\small
\Require traffic simulator $\rho_{\mathbf{S}}(\mathbf{s}_0, \bm{\epsilon})$ which performs a simulation with the specified initial vehicle states $\mathbf{s}_0$ and temporal observation error sequence $\bm{\epsilon}$ and returns the robustness of the simulation, where $\mathbf{S} \in \{\text{east}, \text{west}, \text{south}, \text{north}\}$ is the scenario
\Require initial denoising diffusion model $p_{\theta_0}(\bm{\epsilon} \mid \rho_{\text{threshold}}, \mathbf{s}_0)$
\Require prior sensor noise (observation error) model $p(\bm{\epsilon})$
\Require training batch size $N$
\Function{Trainer}{$p_{\theta_0}(\bm{\epsilon} \mid \rho_{\text{threshold}}, \mathbf{s}_0), \rho_{\mathbf{S}}(\mathbf{s}_0, \bm{\epsilon}), p(\bm{\epsilon}), N$}
    \State sample initial traffic states in scenario $\{\mathbf{s}_0^{(n)} \sim \mathbf{S}\}_{n=1}^N$
    \State sample observation error from prior $\{\bm{\epsilon}^{(n)} \sim p(\bm{\epsilon})\}_{n=1}^{N}$
    \State evaluate robustness $\{\rho^{(n)} \gets \rho_{\mathbf{S}}(\mathbf{s}_0^{(n)}, \bm{\epsilon}^{(n)})\}_{n=1}^N$
    \State update elite cutoff $\Tilde{\rho} \gets \text{quantile}(\alpha, \{\rho^{(n)}\}_{n=1}^N)$
    \State create dataset $\mathcal{D} \gets \{(\bm{\epsilon}^{(n)}, \rho^{(n)}, \mathbf{s}_0^{(n)})\}_{n=1}^N$
    \State $\theta \gets \text{TRAIN}(\theta_0, \mathcal{D})$
    \While{$\Tilde{\rho}$ has not converged}
        \State resample initial traffic states $\{\mathbf{s}_0^{(n)} \sim \mathbf{S}\}_{n=1}^N$
        \State sample robustness thresholds $\{\rho_{\text{threshold}}^{(n)} \sim \mathcal{U}[0, \Tilde{\rho}]\}_{n=1}^N$
        \State sample $\{\bm{\epsilon}^{(n)} \sim p_{\theta}(\bm{\epsilon} \mid \rho_{\text{threshold}}^{(n)}, \textbf{s}_0^{(n)})\}_{n=1}^N$
        \State evaluate robustness $\{\rho^{(n)} \gets \rho_{\mathbf{S}}(\mathbf{s}_0^{(n)}, \bm{\epsilon}^{(n)})\}_{n=1}^N$
        \State update elite cutoff $\Tilde{\rho} \gets \text{quantile}(\alpha, \{\rho^{(n)}\}_{n=1}^N)$
        \State $\mathcal{D} \gets \mathcal{D} \cup \{(\bm{\epsilon}^{(n)}, \rho^{(n)}, \mathbf{s}_0^{(n)})\}_{n=1}^N$
        \State $\theta \gets \text{TRAIN}(\theta, \{(\bm{\epsilon}, \rho, \mathbf{s}_0) \in \mathcal{D} \mid \rho \leq \Tilde{\rho}\})$
    \EndWhile
    \State \Return $\theta$
\EndFunction
\end{algorithmic}
\end{algorithm}

We train a separate diffusion model for each of the four scenarios because the scenarios are fundamentally different and thus likely have very different failure distributions. Furthermore, the algorithm keeps a running dataset that is continuously replenished by newly generated data; this helps ensure a sufficient number of elite training samples even when the elite robustness cutoff $\Tilde{\rho}$ becomes very low in the later stages of training.

\subsection{Evaluation Metrics and Baselines}
We aim to evaluate the extent to which the simulations generated by the diffusion model match the true distribution of potential failures in the scenario of interest. By ``simulations generated by the diffusion model," we refer to the simulations (i.e. vehicle state trajectories) that occur as a result of the observation error sequences generated by the diffusion model. We propose the following evaluation metrics:

\begin{itemize}
    \item \textbf{Failure Rate}: Fraction of simulations generated that are actual collisions. A high failure rate indicates less waste of computation in generating non-collision samples, i.e. a sample-efficient generation of failures. This metric is inclusively bounded between 0 and 1.
    \item \textbf{Density}: The extent to which the simulations generated tend to match the prevalent failure modes in the true failure distribution. A higher density indicates that the model tends to generate more realistic failures. This metric is unbounded and can exceed 1 \cite{naeem2020reliable}.
    \item \textbf{Coverage}: The extent to which the simulations generated capture the full range of variability of the true failure distribution. A high coverage means the model can generate a greater variety of different failures. This metric is inclusively bounded between 0 and 1 \cite{naeem2020reliable}.
\end{itemize}

We compare our method against the cross-entropy method (CEM) \cite{rubinstein2016simulation}. This algorithm iteratively updates a Gaussian model to minimize its KL divergence to the true failure distribution, which is estimated through Monte Carlo Simulations.

\section{Experiments and results} \label{Experiments}
\subsection{Environment Setup and Monte Carlo Simulations}
Each lane in the HighwayEnv simulation environment is 0.04 units wide. We configure the environments such that the ego vehicle's initial distance to the intersection is sampled from $\mathcal{U}[0.35, 0.65]$ and its initial velocity from $\mathcal{U}[0.35, 0.5]$ (in units of length per second). The intruder's initial distance to the intersection is sampled from $\mathcal{U}[0.25, 0.45]$ and its speed from $\mathcal{U}[0.35, 0.45]$. The respective destinations of the vehicles and the behavior of the intruder are randomized based on HighwayEnv's provided feature. To ensure a reasonable failure rate, the sensor noise model of the ego vehicle is set to $\mathcal{N}(\mathbf{0}, 1/0.15\mathbf{I})$. To estimate the true failure distribution of the system in the four scenarios, we performed a large number of Monte Carlo simulations on a dedicated computer cluster over a period of 3 weeks. The results are in \cref{mc_results}.
\vspace{-0.3cm}
\begin{table}[htbp]
\caption{Monte Carlo Simulation Results}
% \vspace{-0.7cm}
\label{mc_results}
\begin{center}
\begin{adjustbox}{max width=\columnwidth}
\begin{tabular}{@{}lrrr@{}}
\toprule
\textbf{Scenario} & \textbf{Simulations Performed} & \textbf{Failures Found} & \textbf{Failure Rate} \\
\midrule
South & \num{707200000} & \num{1028} & \num{1.45e-6}\\
North & \num{2500000} & \num{1892} & \num{7.57e-4} \\
West & \num{50000000} & \num{1083} & \num{2.17e-05} \\
East & \num{11000000} & \num{1017} & \num{9.25e-05} \\
% \hline
% \hline
% \hline
\bottomrule
\end{tabular}
\end{adjustbox}
\label{tab1}
\end{center}
\end{table}

\vspace{-0.65cm}
\subsection{Diffusion Model Training and Evaluation}
Our diffusion model is based on a U-Net with four downsampling and upsampling layers, which we implemented using PyTorch. We set the elite sample cutoff ratio $\alpha$ to 0.1. A separate model is trained for each scenario on a GeForce GTX 1080Ti GPU using the AdamW optimizer with a learning rate of $\num{3e-4}$ and a batch size ($N$ as defined in \cref{alg:diffusion_failure_sampling}) of 256. The model took about 5 iterations (16 hours) to converge for the south scenario, 10 iterations (31 hours) for west, 11 iterations (33 hours) for east, and 30 iterations (90 hours) for north. We then drew over 1000 failure samples from the model for each scenario and compared the diffusion samples against the Monte Carlo failure samples to evaluate the sample fidelity of the models. The CEM algorithm required roughly 32 hours of runtime for each scenario, and it was evaluated using the same method. The evaluation results are shown in \cref{results}.

% \begin{table}[h!]
%     \centering
%     \caption{Performance metrics for the diffusion model and the Cross-Entropy Method (CEM) in all four scenarios.}
%     \begin{tabular}{@{}ccccc@{}}
%         \toprule
%         \textbf{Scenario} & \textbf{Method} & \textbf{failure rate  $\uparrow$} & \textbf{density $\uparrow$} & \textbf{coverage $\uparrow$} \\
%         \midrule
%         East   & diffusion model & \textbf{0.0160} & \textbf{0.8924} & \textbf{0.7837} \\
%         % \cline{2-5}
%                & CEM             & 0.0138 & 0.6387 & 0.3805 \\
%         \midrule
%         West   & diffusion model & \textbf{0.0094} & \textbf{0.8676} & \textbf{0.4314} \\
%         % \cline{2-5}
%                & CEM             & 0.0069 & 0.7678 & 0.2410 \\
%         \midrule
%         North  & diffusion model & 0.0048 & \textbf{0.9558} & \textbf{0.8967} \\
%         % \cline{2-5}
%                & CEM             & \textbf{0.0130} & 0.6677 & 0.4752 \\
%         \midrule
%         South   & diffusion model & \textbf{0.2379} & \textbf{0.9451} & \textbf{0.8122} \\
%         % \cline{2-5}
%                & CEM             & 0.0 & / & / \\
%         \bottomrule
%     \end{tabular}
%     \label{results}
% \end{table}

\begin{table}[ht]
    \centering    \caption{Performance metrics for the diffusion model and the Cross-Entropy Method (CEM) in all four scenarios.}
    \sisetup{detect-weight=true, detect-family=true}
    \begin{adjustbox}{max width=\columnwidth}
    \begin{tabular}{@{}llS[table-format=2.3(4), separate-uncertainty=true, retain-zero-uncertainty = true]S[table-format=2.3(4), separate-uncertainty=true, retain-zero-uncertainty = true]S[table-format=2.3(4), separate-uncertainty=true, retain-zero-uncertainty = true]@{}}
        \toprule
        \textbf{Scenario} & \textbf{Method} & \textbf{Failure Rate $\uparrow$} & \textbf{Density $\uparrow$} & \textbf{Coverage $\uparrow$} \\
        \midrule
        \multirow{2}{*}{\parbox{1cm}{South}}  & Diffusion model & \textbf{\num{0.2134}} & \textbf{\num{0.9002}} & \textbf{\num{0.8045}} \\
               & CEM             & \num{0.0000} & / & / \\
        \midrule
        \multirow{2}{*}{\parbox{1cm}{North}}   & Diffusion model & \num{0.0048} & \textbf{\num{0.9558}} & \textbf{\num{0.8967}} \\
               & CEM             & \textbf{\num{0.0130}} & \num{0.6677} & \num{0.4752} \\
        \midrule
        \multirow{2}{*}{\parbox{1cm}{West}}    & Diffusion model & \textbf{\num{0.0094}} & \textbf{\num{0.8676}} & \textbf{\num{0.4314}} \\
               & CEM             & \num{0.0069} & \num{0.7678} & \num{0.2410} \\
        \midrule
        \multirow{2}{*}{\parbox{1cm}{East}}   & Diffusion model & \textbf{\num{0.0160}} & \textbf{\num{0.8924}} & \textbf{\num{0.7837}} \\
               & CEM             & \num{0.0138} & \num{0.6387} & \num{0.3805} \\
        \bottomrule
    \end{tabular}
    \end{adjustbox}
    \label{results}
\end{table}

% \begin{figure*}
%   % Include PDFs here
%   \caption{Visualization of failure cases obtained from Monte Carlo simulations, diffusion model sampling, and CEM sampling. Failure cases are represented as failure trajectories, which are temporal series of the intruder's position relative to the ego.} 
%   \label{results_viz}
% \end{figure*}

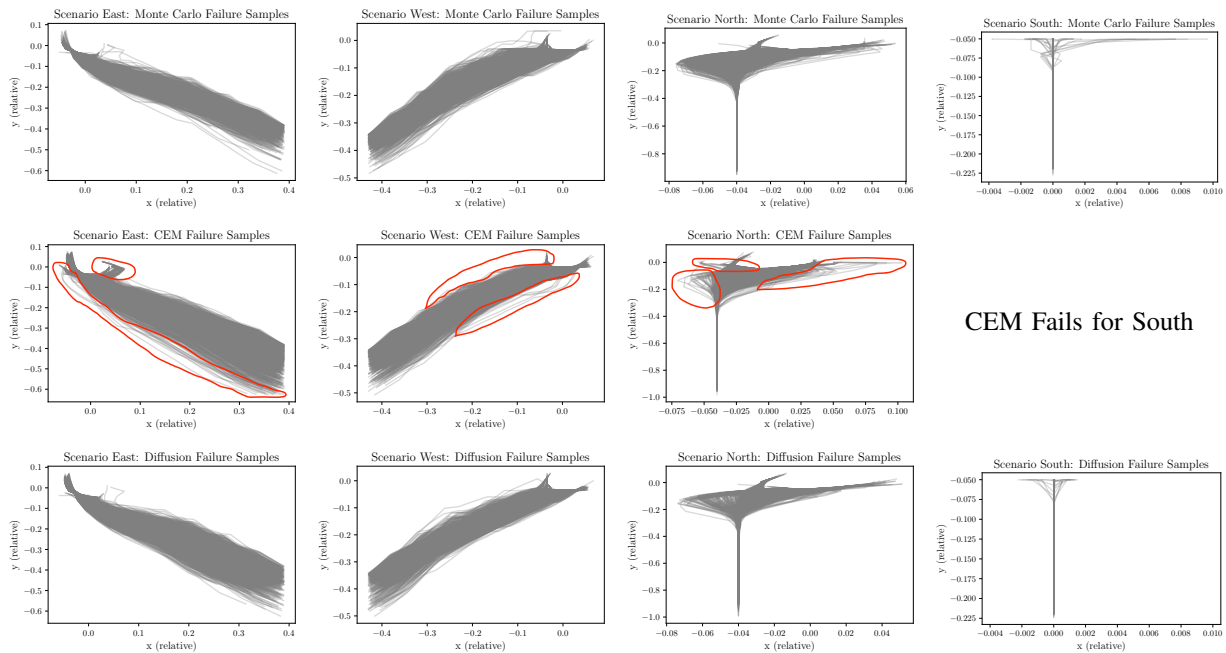
\begin{figure*}
    \centering
    \setlength{\tabcolsep}{2pt} % Adjust horizontal spacing
    \renewcommand{\arraystretch}{1} % Adjust row spacing

    \begin{tabular}{c c c c}  
        \includegraphics[width=0.22\textwidth]{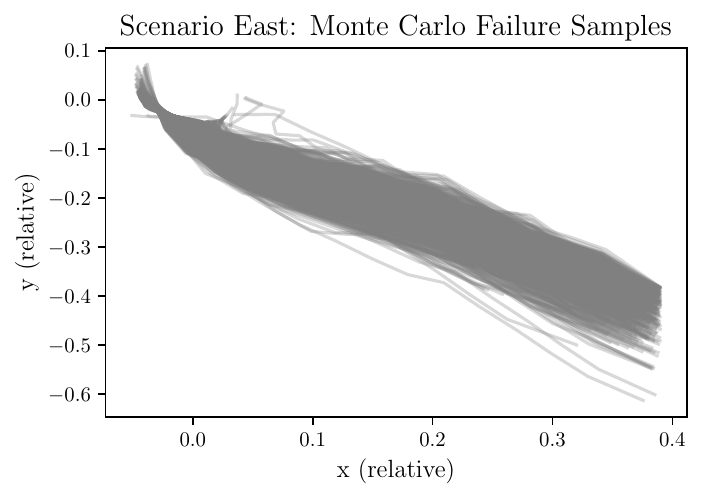} &  
        \includegraphics[width=0.22\textwidth]{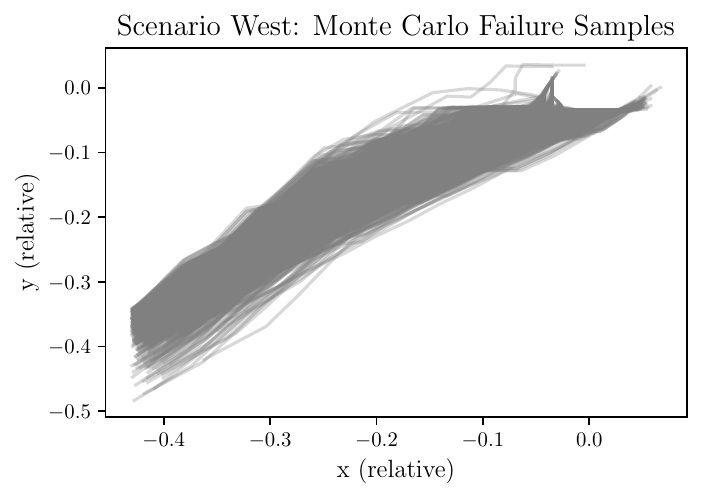} &  
        \includegraphics[width=0.22\textwidth]{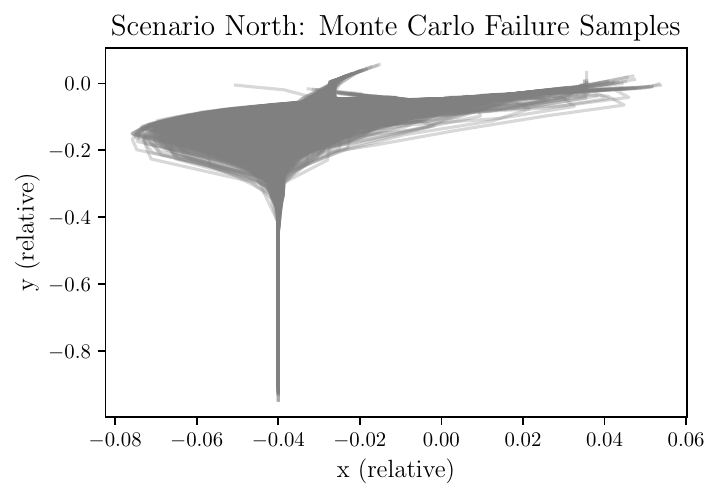} &  
        \includegraphics[width=0.22\textwidth]{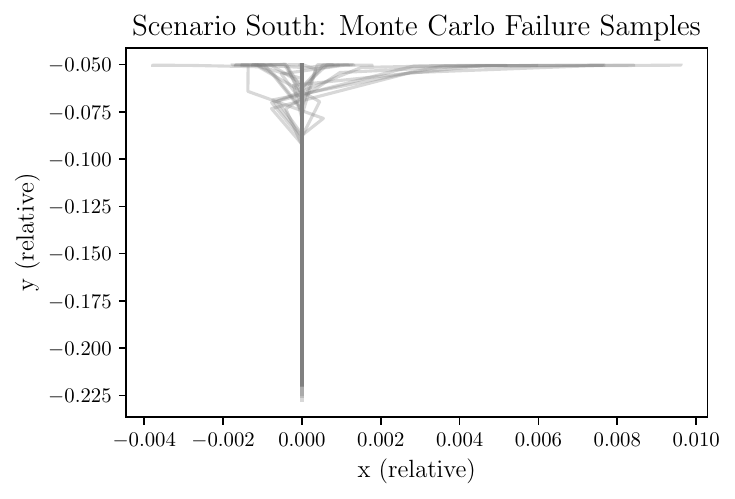} \\

        \includegraphics[width=0.22\textwidth]{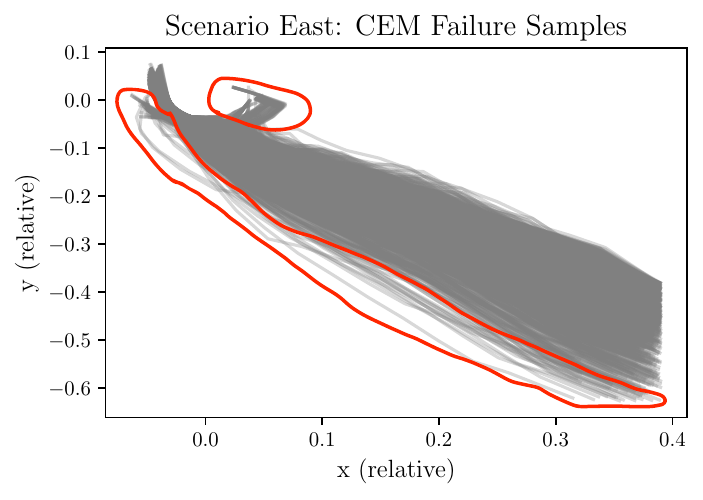} &  
        \includegraphics[width=0.22\textwidth]{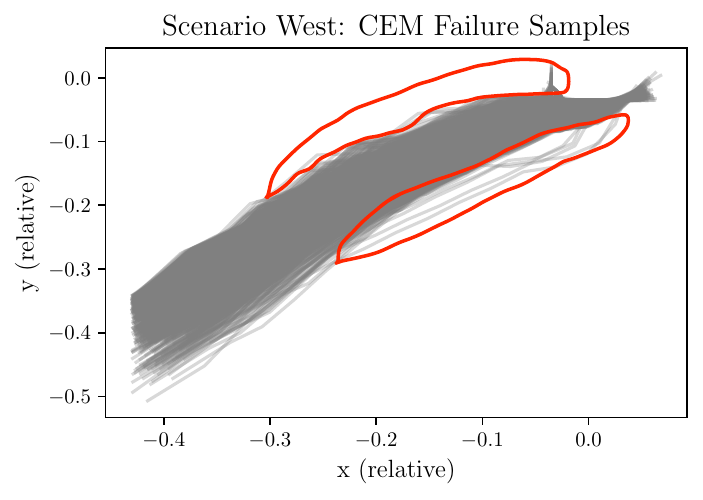} &  
        \includegraphics[width=0.22\textwidth]{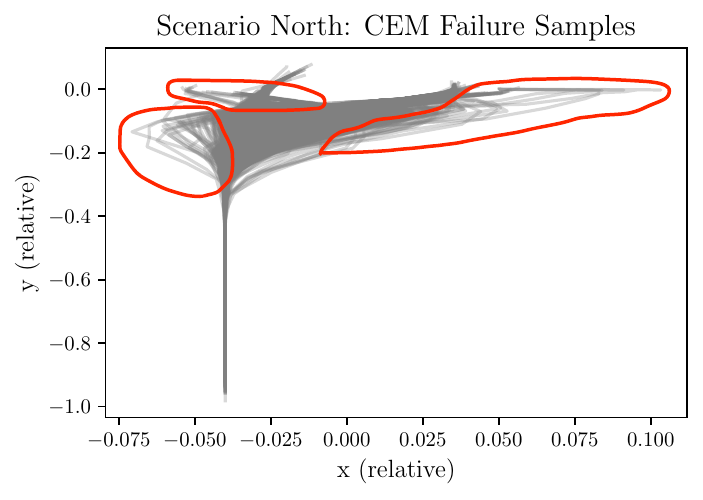} &  \raisebox{1.3cm}{ \begin{tikzpicture} \node at (0,0) {CEM Fails for South}; \end{tikzpicture} } \\

        \includegraphics[width=0.22\textwidth]{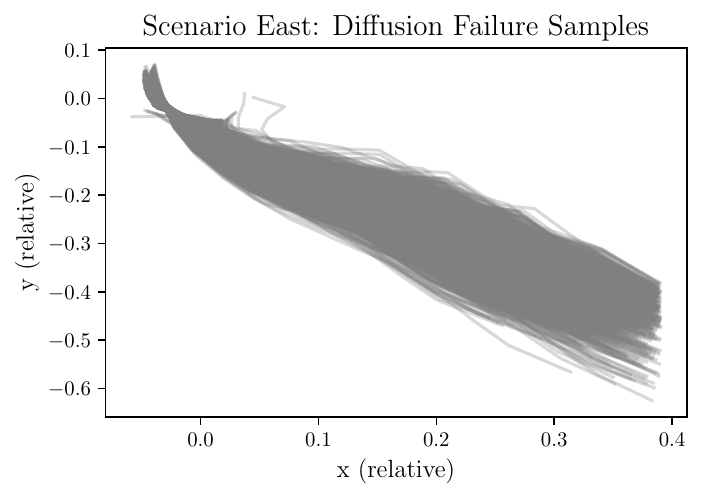} &  
        \includegraphics[width=0.22\textwidth]{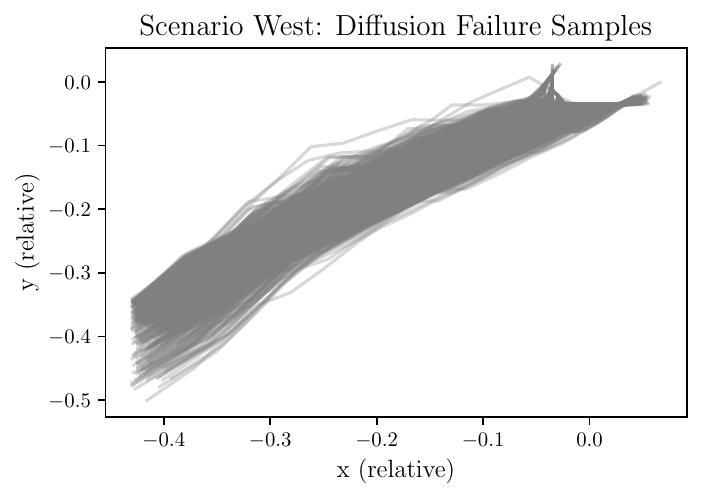} &  
        \includegraphics[width=0.22\textwidth]{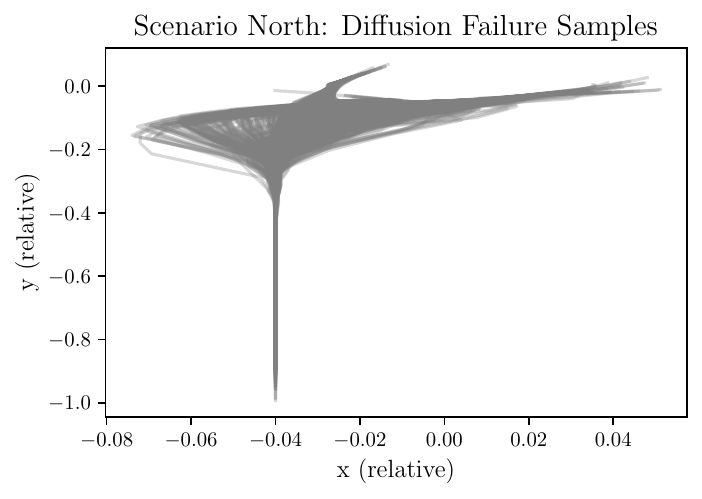} &  
        \includegraphics[width=0.22\textwidth]{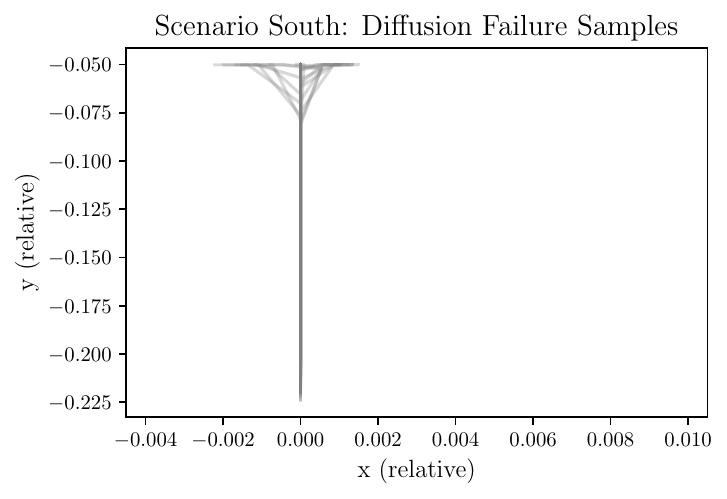} \\
    \end{tabular}

    \caption{Visualization of failure cases obtained from Monte Carlo simulations, diffusion model sampling, and CEM sampling. Failure cases are represented as failure trajectories, which are temporal series of the intruder's relative position.} 
    \label{results_viz}
\end{figure*}

\section{Discussions} \label{Discussions}
Our first observation is that in the south scenario, illustrated in \cref{problem viz}, the CEM algorithm is unable to generate any failure samples. In contrast, the diffusion model achieves a failure rate as high as 21.34\%, showcasing its efficiency in generating sensor noise samples that lead to actual collisions. This scenario, as shown in \cref{mc_results}, features an extremely low probability of collision under a Gaussian sensor noise model. Furthermore, \cref{results_viz} shows that the true (Monte Carlo) failure distribution for scenario south is highly dispersed without any apparent prominent failure mode. While both the diffusion model and CEM train themselves to progressively generate samples that are increasingly close to collisions, CEM assumes the Gaussian nature of the failure space, which significantly limits its capacity to represent diffuse failure distributions as observed in the south scenario. The diffusion model, however, learns to transform a Gaussian distribution to the true failure distribution through multiple U-Net denoising steps, which makes it able to effectively capture the failure space even if the failure distribution is highly complex. A few failure trajectories not captured by the diffusion model involve unusual horizontal displacement of the vehicles: these trajectories are caused by abnormal post-collision steering behavior of the vehicles and are very rare events, and thus the diffusion model still achieves very high coverage (0.80).

Across the other three scenarios, the diffusion model outperforms CEM in both density and coverage, demonstrating superior sample fidelity. As shown in \cref{results_viz}, for the east scenario, CEM significantly oversamples failures in the two circled regions. These failures are rarely represented in the true failure distribution, suggesting the tendency of the CEM algorithm to over-generate unrealistic or unlikely failures. The diffusion model's sample distribution, however, matches the true failure distribution better, as reflected in its much higher density metric compared with CEM (0.89 versus 0.64). We also notice that for the west scenario, the CEM algorithm is ineffective in capturing failure trajectories in the two circled regions which are otherwise featured prominently in the true failure distribution. The diffusion model captures failures in these two regions significantly better, resulting in its much higher coverage score (0.43 versus 0.24). For the north scenario, the CEM algorithm noticeably undersamples failures in the far-left and far-right circled areas compared with the true failure distribution; it also significantly oversamples failures in the middle circle, which is a relatively rare failure mode based on the true distribution. On the other hand, the sample distribution of the diffusion model in these regions matches that of the true distribution substantially better, as shown in its much higher density (0.96 versus 0.67) and coverage (0.90 versus 0.48) scores. Notably, the CEM method performs worse on coverage in each scenario: this may be due to the inherent mode-seeking behavior of the CEM algorithm, which stems from the Gaussian assumption of its proposal distribution.

Finally, the diffusion model achieves a higher failure rate than CEM in all but the north scenario, showcasing a generally higher efficiency in failure generation. There is occasionally a tradeoff between failure rate and the other metrics (especially coverage). In order for a generative model to capture a larger span of the true failure distribution, it cannot simply focus on generating one or a few most likely failure modes. When generating diverse observation error samples, some of which may be less likely to cause collisions, the model unavoidably compromises its failure rate and yet gains coverage over more potential types of failure. Conversely, a high failure rate sometimes entails compromised coverage, which we observe in the mode-collapse behavior of CEM for the north scenario, as shown in \cref{results_viz}.

\section{Conclusion}
We propose a framework for safety validation of autonomous vehicles using generative artificial intelligence. Focusing on the traffic intersection problem, we train a diffusion model to generate temporal series of sensor noise that may cause an autonomous vehicle to collide with an intruder based on any specified initial traffic state. In all scenarios tested, failure samples from the trained models exhibit high density and coverage, demonstrating the ability of the models to generate realistic failure cases based on the initial state while capturing a wide variety of possible failures of the SUT. The diffusion model achieves higher failure rate than the CEM algorithm in the vast majority of scenarios, demonstrating a sample-efficient generation of failures.

We envision that our framework may also be adapted to robust planning applications in autonomous driving. Future work can leverage techniques such as knowledge distillation \cite{sauer2025adversarial} to compress our models and deploy the distillation models on hardware to support real-time generation of potential failures based on current traffic states. Such data provide valuable information for onboard active collision avoidance systems and may substantially improve the operational safety of autonomous vehicles.

\bibliographystyle{ieeetr}
\bibliography{references}

\begin{thebibliography}{10}

\bibitem{junietz2018evaluation}
P.~Junietz, W.~Wachenfeld, K.~Klonecki, and H.~Winner, ``Evaluation of different approaches to address safety validation of automated driving,'' in {\em IEEE International Conference on Intelligent Transportation Systems (ITSC)}, pp.~491--496, IEEE, 2018.

\bibitem{dona2022virtual}
R.~Don{\`a} and B.~Ciuffo, ``Virtual testing of automated driving systems. {A} survey on validation methods,'' {\em IEEE Access}, vol.~10, pp.~24349--24367, 2022.

\bibitem{saharia2022photorealistic}
C.~Saharia, W.~Chan, S.~Saxena, L.~Li, J.~Whang, E.~L. Denton, K.~Ghasemipour, R.~Gontijo~Lopes, B.~Karagol~Ayan, T.~Salimans, {\em et~al.}, ``Photorealistic text-to-image diffusion models with deep language understanding,'' in {\em Advances in Neural Information Processing Systems (NeurIPS)}, pp.~36479--36494, 2022.

\bibitem{rombach2022high}
R.~Rombach, A.~Blattmann, D.~Lorenz, P.~Esser, and B.~Ommer, ``High-resolution image synthesis with latent diffusion models,'' in {\em IEEE Computer Society Conference on Computer Vision and Pattern Recognition (CVPR)}, pp.~10684--10695, 2022.

\bibitem{janner2022planning}
M.~Janner, Y.~Du, J.~B. Tenenbaum, and S.~Levine, ``Planning with diffusion for flexible behavior synthesis,'' in {\em International Conference on Machine Learning (ICML)}, 2022.

\bibitem{li2023simulation}
C.~Li, J.~Sifakis, Q.~Wang, R.~Yan, and J.~Zhang, ``Simulation-based validation for autonomous driving systems,'' in {\em ACM SIGSOFT International Symposium on Software Testing and Analysis}, pp.~842--853, 2023.

\bibitem{tlig2018autonomous}
M.~Tlig, M.~Machin, R.~Kerneis, E.~Arbaretier, L.~Zhao, F.~Meurville, and J.~Van~Frank, ``Autonomous driving system: Model based safety analysis,'' in {\em IEEE/IFIP International Conference on Dependable Systems and Networks Workshops (DSN-W)}, pp.~2--5, 2018.

\bibitem{jiang2022efficient}
Z.~Jiang, W.~Pan, J.~Liu, S.~Dang, Z.~Yang, H.~Li, and Y.~Pan, ``Efficient and unbiased safety test for autonomous driving systems,'' {\em IEEE Transactions on Intelligent Vehicles}, vol.~8, no.~5, pp.~3336--3348, 2022.

\bibitem{xu2019quantitative}
B.~Xu, Q.~Li, T.~Guo, Y.~Ao, and D.~Du, ``A quantitative safety verification approach for the decision-making process of autonomous driving,'' in {\em International Symposium on Theoretical Aspects of Software Engineering (TASE)}, pp.~128--135, 2019.

\bibitem{corso2020interpretable}
A.~Corso and M.~J. Kochenderfer, ``Interpretable safety validation for autonomous vehicles,'' in {\em IEEE International Conference on Intelligent Transportation Systems (ITSC)}, pp.~1--6, 2020.

\bibitem{feng2023dense}
S.~Feng, H.~Sun, X.~Yan, H.~Zhu, Z.~Zou, S.~Shen, and H.~X. Liu, ``Dense reinforcement learning for safety validation of autonomous vehicles,'' {\em Nature}, vol.~615, no.~7953, pp.~620--627, 2023.

\bibitem{koren2018adaptive}
M.~Koren, S.~Alsaif, R.~Lee, and M.~J. Kochenderfer, ``Adaptive stress testing for autonomous vehicles,'' in {\em IEEE Intelligent Vehicles Symposium (IV)}, 2018.

\bibitem{ho2020denoising}
J.~Ho, A.~Jain, and P.~Abbeel, ``Denoising diffusion probabilistic models,'' in {\em Advances in Neural Information Processing Systems (NeurIPS)}, pp.~6840--6851, 2020.

\bibitem{galib2024fide}
A.~H. Galib, P.-N. Tan, and L.~Luo, ``Fide: Frequency-inflated conditional diffusion model for extreme-aware time series generation,'' in {\em Advances in Neural Information Processing Systems (NeurIPS)}, 2024.

\bibitem{chen2023advdiffuser}
X.~Chen, X.~Gao, J.~Zhao, K.~Ye, and C.-Z. Xu, ``Advdiffuser: Natural adversarial example synthesis with diffusion models,'' in {\em International Conference on Computer Vision (ICCV)}, pp.~4562--4572, 2023.

\bibitem{guo2024drivinggen}
Z.~Guo, Y.~Zhou, and C.~Gou, ``Drivinggen: Efficient safety-critical driving video generation with latent diffusion models,'' in {\em IEEE International Conference on Multimedia and Expo (ICME)}, pp.~1--6, 2024.

\bibitem{chang2025safe}
W.-J. Chang, F.~Pittaluga, M.~Tomizuka, W.~Zhan, and M.~Chandraker, ``Safe-sim: Safety-critical closed-loop traffic simulation with diffusion-controllable adversaries,'' in {\em European Conference on Computer Vision}, pp.~242--258, 2025.

\bibitem{delecki2024diffusion}
H.~Delecki, M.~R. Schlichting, M.~Arief, A.~Corso, M.~Vazquez-Chanlatte, and M.~J. Kochenderfer, ``Diffusion-based failure sampling for evaluating safety-critical autonomous systems,'' {\em arXiv preprint arXiv:2406.14761}, 2024.

\bibitem{kesting2010enhanced}
A.~Kesting, M.~Treiber, and D.~Helbing, ``Enhanced intelligent driver model to access the impact of driving strategies on traffic capacity,'' {\em Philosophical Transactions of the Royal Society A: Mathematical, Physical and Engineering Sciences}, vol.~368, no.~1928, pp.~4585--4605, 2010.

\bibitem{ronneberger2015u}
O.~Ronneberger, P.~Fischer, and T.~Brox, ``U-net: Convolutional networks for biomedical image segmentation,'' in {\em International Conference on Medical Image Computing and Computer-Assisted Intervention (MICCAI)}, pp.~234--241, Springer, 2015.

\bibitem{naeem2020reliable}
M.~F. Naeem, S.~J. Oh, Y.~Uh, Y.~Choi, and J.~Yoo, ``Reliable fidelity and diversity metrics for generative models,'' in {\em International Conference on Machine Learning (ICML)}, pp.~7176--7185, 2020.

\bibitem{rubinstein2016simulation}
R.~Y. Rubinstein and D.~P. Kroese, {\em Simulation and the Monte Carlo method}.
\newblock John Wiley \& Sons, 2016.

\bibitem{sauer2025adversarial}
A.~Sauer, D.~Lorenz, A.~Blattmann, and R.~Rombach, ``Adversarial diffusion distillation,'' in {\em European Conference on Computer Vision}, pp.~87--103, Springer, 2025.

\end{thebibliography}

\end{document}